\documentclass[ejsv2,preprint,noshowframe]{imsart}
\IfFileExists{newtxmath.sty}{}{\RequirePackage{amssymb}}
\RequirePackage[numbers,sort&compress]{natbib}
\RequirePackage[hidelinks,hypertexnames=false]{hyperref}
\RequirePackage{xcolor}
\RequirePackage{microtype}

\startlocaldefs
\newcommand{\X}{\mathcal X}
\newcommand{\D}{\mathcal D}
\newcommand{\EXP}{\mathbb E}
\newcommand{\PROB}{\mathbb P}
\newcommand{\IND}{\mathbf 1}

\theoremstyle{plain}
\newtheorem{theorem}{Theorem}
\newtheorem{proposition}{Proposition}
\newtheorem{corollary}{Corollary}
\newtheorem{lemma}{Lemma}
\theoremstyle{definition}
\newtheorem{definition}{Definition}
\theoremstyle{definition}
\newtheorem{remark}{Remark}
\allowdisplaybreaks[2]

\endlocaldefs

\begin{document}
\begin{frontmatter}
\pdfsubject{}
\title{On \protect\texorpdfstring{$k$}{k}-nearest-neighbor regression in metric spaces}
\runtitle{$k$-NN regression in metric spaces}

\begin{aug}
\author{\fnms{L\'aszl\'o}~\snm{Gy\"orfi}},
\author{\fnms{Vladimir G.}~\snm{Pestov}},
\author{\fnms{Harro}~\snm{Walk}}
\and
\author{\fnms{Roi}~\snm{Weiss}}
\runauthor{L. Gy\"orfi et al.}
\end{aug}

\begin{abstract}
We study mean squared consistency of $k$-nearest-neighbor
regression for metric space valued features. 
For a fixed feature distribution $\mu$ with atomless distance distributions, we describe how neighbor
selection redistributes
probability through the density of a uniformly chosen neighbor of
an independent query. This gives separate conditions for the
approximation error and the conditional response variance, and
weakens Stone's domination condition when stronger response moments
are available. In particular, when $k_n\to\infty$ and
$k_n/n\to0$, a uniform $L_2(\mu)$ bound on the selection densities
suffices under a finite fourth response moment. 
We give a corresponding condition in terms of neighborhoods
of fixed population mass and recover the Euclidean density bound.
Finite global Nagata dimension gives Stone's domination.
While sigma-finite Nagata dimensionality is known to ensure
universal consistency of $k$-NN classification in complete separable
spaces, we show that this is not the case for mean squared
consistency with unbounded responses: we construct a space in this
class on which one fixed noiseless response with all finite moments
has diverging mean squared error.
\end{abstract}



\end{frontmatter}

\section{Introduction}
\label{sec:introduction}
\label{sec:setup}
Let $(\X,\rho)$ be a separable metric space with its Borel
$\sigma$-algebra.
For the estimation of the nonparametric regression function, let the random feature (predictor) $X$ take values in the space $\X$, and let its response (label)
$Y$ be real valued.
We denote by $\mu$ the feature distribution, that is, $\mu(A)=\PROB (X\in A)$ for all measurable sets $A\subseteq \X$.
For the mean squared error analysis, assume $\EXP Y^2<\infty$.
Define measurable versions of the regression function and the
conditional variance by
\begin{align}
\label{def:model-m}
m(x)&=\EXP[Y\mid X=x],\\[4pt]
\label{def:model-sigma}
\sigma^2(x)&=\EXP[(Y-m(X))^2\mid X=x].
\end{align}
Let $\D_n=\{(X_1,Y_1),\ldots,(X_n,Y_n)\}$ consist of independent
copies of $(X,Y)$. A test pair, again denoted by $(X,Y)$, is
independent of $\D_n$. For a query $x$, order the observations $X_i$ by
increasing distance  $\rho(x,X_i)$ and denote them by
\[
(X_{(n,1)}(x),Y_{(n,1)}(x)),\ldots,
(X_{(n,n)}(x),Y_{(n,n)}(x)).
\]
Thus $X_{(n,k)}(x)$ is the $k$-th nearest neighbor of $x$.
Ties are resolved by indices. For integers $1\le k_n<n$, $n\ge2$,
satisfying
\begin{equation}
\label{nkn}
k_n\longrightarrow\infty,\qquad k_n/n\longrightarrow0,
\end{equation}
the estimate and its mean integrated squared error are
\begin{align}
\label{estimator}
m_n(x)&=\frac1{k_n}\sum_{i=1}^{k_n}Y_{(n,i)}(x),\\[3pt]
\label{risk}
R_{n,2}&=\EXP\int(m_n(x)-m(x))^2\,\mu(dx).
\end{align}
The relevance of this criterion comes from the mean squared error identity
\[
\EXP[(Y-m_n(X))^2]=\EXP[\sigma^2(X)]+R_{n,2}.
\]
The first term is the irreducible prediction error. Thus
$R_{n,2}\to0$ means that the mean squared prediction error approaches
the expected conditional variance, $\EXP[\sigma^2(X)]$. We call this \emph{weak consistency in
$L_2(\mu)$}. 
Background on nearest-neighbor regression is given in
\citet[Chapter~6]{GyKoKrWa02} and \citet[Chapter~11]{BiDe15}.

Separability of the metric space $\X$ alone does not ensure consistency of ordinary $k$-NN.
There are feature distributions on the infinite-dimensional Hilbert
space $\ell_2$, with deterministic binary responses, for which the
rule is inconsistent for every sequence satisfying \eqref{nkn};
see \citet[Section~3.2]{cerou2006nearest} and
\citet[Section~3]{CoKuPe20}. Thus locality of the selected
neighbors is not by itself sufficient for estimating an arbitrary
measurable conditional mean.

The failure of ordinary $k$-NN in some separable metric spaces is a
property of this estimator, rather than an impossibility of learning
in those spaces. For classification, the OptiNet rule of
\citet{HaKoSaWe21} is universally strongly Bayes consistent in every
separable metric space. The simpler Proto-NN construction of
\citet{GyWe21} uses an independent unlabeled sample (prototypes) to form a Voronoi
partition of $\X$ and averages the labeled responses within its cells.
If the number $M_n$ of prototypes tends to infinity and $M_n/n\to0$,
their Theorem~3 gives strong integrated consistency of regression for bounded
responses. For unbounded responses, set the estimate to zero in
cells with fewer than $\log n$ observations and require
$M_n\log n/n\to0$. Their Theorem~5 and the subsequent discussion
then give strong $L_2$ consistency whenever $\EXP Y^2<\infty$.

The partition structure explains an important difference. Conditional
on a partition independent of the labeled sample, the expected
cell-average of a nonnegative integrable function $\varphi$ equals
its conditional mean in that cell, multiplied by the probability
that the cell is occupied. Averaging over the independent query
therefore gives
\[
\EXP\sum_{i=1}^n V_{n,i}(X)\varphi(X_i)
\le\int\varphi\,d\mu,
\]
where $V_{n,i}$ are the cell-average weights and are zero on empty
cells. The modified rule for unbounded regression only reduces these weights, leaving the inequality intact. This domination
holds without any dimension bound. Ordinary $k$-NN uses a different
neighborhood for every query, so a separate argument is needed to
control the distribution of the observations it selects.

The classical connection with differentiation of measures also has a
statistical interpretation. Write $B(x,t)$ for the closed ball of
radius $t$ around $x$. For a query $x$ and a radius $t$ with
$\mu(B(x,t))>0$, the population mean response in the ball is
\[
a_t(x):=\EXP[Y\mid X\in B(x,t)]
=\frac{\int_{B(x,t)}m(z)\,\mu(dz)}{\mu(B(x,t))}.
\]
This is the population quantity estimated by a local average over
that ball. Recovering $m(x)$ as $t\downarrow0$ is therefore the
approximation requirement behind local averaging. In Euclidean
spaces the differentiation theorem supplies this requirement almost
everywhere for every Borel probability measure and every integrable
regression function \citep[Theorem~20.19]{BiDe15}. In general metric spaces
it can fail, even for an indicator function. The relevance to
nearest-neighbor classification was developed by
\citet{cerou2006nearest}. Their corrected sufficient condition uses
almost-everywhere differentiation even for weak consistency; see
\citet[Section~2.3, Theorem~2.2]{Pes26}.

To see the role of this condition, fix a query $x$ where
$a_t(x)\to m(x)$ as $t\downarrow0$. Given $\epsilon>0$, there is a
radius $t_0(x,\epsilon)>0$ such that
$|a_t(x)-m(x)|<\epsilon$ for every $0<t<t_0(x,\epsilon)$.
This controls the average also at a random positive nearest-neighbor
radius once that radius is sufficiently small. Convergence in
$\mu$-measure for a common radius $t$ says only that most query
points are well approximated at that radius; it does not provide
the same control when the radius is selected separately for each
query. Here ``almost everywhere'' concerns query points under
$\mu$. It is distinct from the almost-sure convergence along
training sequences required by strong consistency.

The obstruction can occur even when $X$ has a Gaussian distribution.
\citet{PrRiTi21} constructed a nondegenerate Gaussian measure $\mu$
on an infinite-dimensional separable Hilbert space $H$ and a Borel
set $M$ of positive measure for which
\[
\lim_{t\downarrow0}\sup_{x\in H}
\frac{\mu(B(x,t)\cap M)}{\mu(B(x,t))}=0.
\]
For the noiseless response $Y=\IND_M(X)$, the population averages
over small balls therefore approach zero even at points where the
conditional mean equals one. A Gaussian distribution for $X$ alone does
not ensure that local averages recover arbitrary measurable
regression functions.

Pointwise estimation and integrated prediction error address
different statistical questions. Pointwise consistency concerns
the estimate at a fixed query, whereas integrated consistency
averages its error over the distribution $\mu$ of an independent
test point $X$.
\citet{Dev82} studies conditions for pointwise convergence of
nearest-neighbor regression estimates. In metric measure spaces,
\citet{CeCo21} connects Lebesgue points with noiseless
nearest-neighbor pointwise estimation and examines the role of
distance ties. Pointwise convergence alone does not ensure
convergence of the mean integrated squared error: large errors on
sets of small feature probability can still make a substantial
contribution to the integral. If $|Y|\le L$ almost surely, then
$|m_n(x)-m(x)|^2\le4L^2$ for $\mu$-almost every $x$, almost surely.
Pointwise almost-sure convergence for $\mu$-almost every $x$ then
allows passage through both integration and expectation by the
bounded convergence theorem. For unbounded responses there is no
such bound, and additional integrability control is needed.

Strong consistency in $L_2(\mu)$ means
\[
\int|m_n-m|^2\,d\mu\longrightarrow0\qquad\text{almost surely}.
\]
Weak consistency requires the expectation of this integrated error,
taken over the training sample, to tend to zero. Strong consistency
requires the integrated error itself to approach zero for almost
every infinite training sample.
For Euclidean predictors and bounded responses,
\citet[Theorem~1 and Remark~7]{DeGyLuKr94} established strong
integrated consistency under $k_n\to\infty$ and $k_n/n\to0$,
using a specified randomized rule for distance ties. Their
Theorem~2 treats integrated absolute error for integrable
responses under the additional condition $k_n/\log n\to\infty$.
For metric spaces, \citet[Theorem~4.1]{KuPe24} proved strong
consistency of the $k$-NN classifier on complete separable
sigma-finite-dimensional spaces in the absence of distance ties,
under \eqref{nkn}.
Here sigma-finite dimensionality means a countable cover by sets
of finite relative Nagata dimension on positive scales; the
geometric definition is given in Section~\ref{sec:nagata}.
Their proofs also use concentration inequalities to control
deviations of integrated error terms from their expectations over
the training sample.
The results of the present paper concern mean integrated squared
error.

Another statistical question is whether the random integrated
error is accurately described by its expectation. Writing
\[
E_n=\int|m_n(x)-m(x)|^2\,\mu(dx),
\qquad \EXP E_n=R_{n,2},
\]
the corresponding relative-stability property is
$E_n/R_{n,2}\to1$ in probability, when $R_{n,2}>0$ eventually.
Relative stability of integrated errors, including nearest-neighbor
regression errors, is studied by \citet{GyScWa02}.

Mean squared consistency of $k$-nearest-neighbor regression is
studied in Section~\ref{sec:consistency}, the underlying geometric
conditions in Section~\ref{sec:nagata}, and rates of convergence
in Section~\ref{sec:rates}. Proofs of the main consistency and
geometric results are deferred to Section~\ref{sec:proofs}.

\subsection{Stone's condition and the role of the feature distribution}
A local averaging estimate has the form
\begin{equation}
\label{weighted-rule}
\widehat m_n(x)=\sum_{i=1}^n W_{n,i}(x)Y_i,
\end{equation}
where the weights are measurable functions of the query and the
feature sample. For probability weights, $W_{n,i}\ge0$ and
$\sum_iW_{n,i}=1$, the following form of Stone's sufficient
criterion isolates the property that will concern us; see
\citet[Theorem~1]{Sto77}.

\begin{theorem}[Stone's sufficient criterion]
\label{consth1}
Let $\mu$ be a Borel probability measure on a separable metric
space, and let $(W_{n,i})$ be feature-dependent probability weights.
Suppose that:
\begin{enumerate}
\item there is $C<\infty$ such that for every $n$ and every
nonnegative integrable Borel function $\varphi$,
\begin{equation}
\tag{S}\label{stone}
\EXP\sum_{i=1}^n W_{n,i}(X)\varphi(X_i)
\le C\int\varphi\,d\mu;
\end{equation}
\item for every $a>0$,
$\sum_iW_{n,i}(X)\IND_{\{\rho(X_i,X)>a\}}\to0$
in probability;
\item $\max_iW_{n,i}(X)\to0$ in probability.
\end{enumerate}
Then $\EXP\int|\widehat m_n-m|^2\,d\mu\to0$ for every joint
law with feature marginal $\mu$ and $\EXP Y^2<\infty$.
\end{theorem}

The usual approximation and truncation proof applies on a separable
metric space: bounded Lipschitz functions approximate square-integrable
regression functions in $L_2(\mu)$, and the three displayed conditions control
approximation, locality, and response noise. Euclidean geometry is
used to verify the domination condition \eqref{stone} for ordinary nearest
neighbors. We use only the sufficient part of the criterion here.

For $k_n$-NN,
\begin{equation}
\label{knn-weights}
W_{n,i}(x)=\frac1{k_n}
\IND_{\{X_i\text{ is among the }k_n\text{ selected neighbors of }x\}}.
\end{equation}
The maximum weight is $1/k_n$. Locality on a separable metric
space gives, under $k_n/n\to0$,
\begin{equation}
\label{locality}
\rho(x,X_{(n,k_n)}(x))\longrightarrow0
\quad\text{almost surely, for $\mu$-almost every }x;
\end{equation}
see \citet[Proposition~2.1]{cerou2006nearest}  and
\citet[Lemma~6.1]{GyKoKrWa02}. For the one-neighbor
locality result, see \citet[Section~V]{CoHa67}. Thus the last two
conditions hold under \eqref{nkn}. In finite-dimensional normed
spaces, a cone-covering argument supplies the remaining domination
bound by controlling how often an observation can be selected as
a neighbor; see \citet[Lemma~4.3 and Remarks~4.4--4.5]{CoKuPe20}.

For functional predictors, a dimension bound of this kind need not
be available. A condition on the feature distribution can still
make the ordinary rule consistent. Our questions are therefore:
how can Stone's domination be weakened when a fourth response
moment, or separate moments of the conditional mean and variance,
are available?
Which geometric assumptions provide these distributional bounds,
and how do they relate to convergence rates?

\section{Neighbor selection and mean squared consistency}
\label{sec:consistency}
\subsection{Neighbor-selection densities}
\label{sec:reverse-density}

To answer these questions, we introduce neighbor-selection
densities $f_n$ on $\X$. Each is the density, relative to $\mu$, of an
observation chosen uniformly among the $k_n$ nearest neighbors
of an independent query $X$. These densities describe how
neighbor selection redistributes the feature distribution
$\mu$, and their norm and integrability bounds control
$L_2(\mu)$ regression consistency under the corresponding
moment assumptions. Cone-covering and Nagata dimension bounds
depend only on the metric; the densities $f_n$ also depend on
$\mu$ and the neighbor sequence $(k_n)$.

Write
\[
B(x,t)=\{z:\rho(x,z)\le t\}
\]
for the closed ball. For the density
analysis from this section onward, we assume that distance
ties have probability zero. More precisely, we assume that
\begin{equation}
\label{defhx}
H_x(t)=\mu(B(x,t)),\qquad t\ge0,
\end{equation}
is continuous on $[0,\infty)$ and satisfies $H_x(0)=0$, for
every $x\in\X$. Thus the distribution of $\rho(x,X)$ has no
atoms, and distance ties have probability zero.

Define the averaging functional, for a
nonnegative Borel function $\varphi$, by
\begin{equation}
\label{An}
A_n(\varphi)=\frac1{k_n}\sum_{i=1}^{k_n}
\EXP\varphi(X_{(n,i)}(X)).
\end{equation}
This is the expectation of $\varphi$ at a point chosen uniformly
among the $k_n$ nearest neighbors of the independent query $X$.
With this notation, Stone's domination condition \eqref{stone}
becomes
\[
A_n(\varphi)\le C\int\varphi\,d\mu
\]
for every $n\ge2$ and every nonnegative $\varphi\in L_1(\mu)$,
with one finite constant $C$ independent of $n$ and $\varphi$.
Selection depends on distances to the query, so the randomly
chosen neighbor may have a different distribution from $\mu$.
Stone's condition compares this distribution with $\mu$.
To study this condition and its relaxations, we describe the
selected-neighbor distribution by a density relative to $\mu$.

For a fixed query $x$ and candidate point $z$, consider the sample
$z,X_2,\ldots,X_n$. The point $z$ is selected precisely when at
most $k_n-1$ of the remaining observations are closer to $x$.
Its selection probability is therefore
\begin{equation}
\label{gn}
g_n(x,z)=\PROB\left\{
  \sum_{i=2}^n\IND_{\{\rho(x,X_i)<\rho(x,z)\}}
  \le k_n-1\right\},
\end{equation}
where the probability is over $X_2,\ldots,X_n$. Averaging over
the query, define the \emph{neighbor-selection density} by
\begin{equation}
\label{fn}
f_n(z)=\frac{n}{k_n}\int g_n(x,z)\,\mu(dx).
\end{equation}
For a fixed query $x$, an observation at $z$ is selected with
probability $g_n(x,z)$. A uniform choice from the $k_n$ selected
observations then chooses that observation with probability
$g_n(x,z)/k_n$. The $n$ sample indices are exchangeable under
\eqref{defhx}, so their equal contributions to the selected-neighbor
distribution sum to give the factor $n/k_n$. The integral over
$x$ in \eqref{fn} averages over the independent query.
These definitions specify measurable functions $g_n$ and $f_n$.

\begin{lemma}[Averaging identity]
\label{averaging-lemma}
For every $n\ge2$ and every nonnegative Borel function $\varphi$,
\begin{equation}
\label{averaging}
A_n(\varphi)=\int\varphi f_n\,d\mu,
\qquad
0\le f_n\le\frac{n}{k_n},
\qquad \int f_n\,d\mu=1.
\end{equation}
In particular, $f_n$ is the density, relative to $\mu$, of a
point chosen uniformly among the $k_n$ nearest neighbors of an
independent query. The averaging identity also holds for every
signed $\varphi\in L_1(\mu)$.
\end{lemma}
The proof is given in Section~\ref{sec:averaging-proofs}.

\subsection{The quadratic decomposition and density conditions}
All norms below are relative to $\mu$. For a measurable function $\phi$ and $1\le p<\infty$,
\[
\|\phi\|_p=\left(\int|\phi|^p\,d\mu\right)^{1/p},
\]
and $\|\phi\|_\infty$ is
the essential supremum. Write $p'$ for the conjugate exponent,
with $1/p+1/p'=1$, $1'=\infty$, and $\infty'=1$. Define
\begin{equation}
\label{density-bounds}
F_p=\sup_{n\ge2}\|f_n\|_p,
\qquad
D_{q,2}=\sup_{n\ge2}\frac{\|f_n\|_q}{k_n},
\qquad 1\le p,q\le\infty.
\end{equation}
By the averaging identity, Stone's condition \eqref{stone} is
equivalent to $F_\infty<\infty$. Indeed, a bound $f_n\le C$
gives the domination inequality. Conversely, applying domination
to $\IND_{\{f_n>C\}}$ shows that $f_n\le C$ almost everywhere.
Each $f_n$ is bounded by $n/k_n$; the issue is control as $n$ varies.

The conditional mean of the estimate given the feature sample is
\begin{equation}
\label{mbar}
\overline m_n(x)=\EXP[m_n(x)\mid X_1,\ldots,X_n]
=\frac1{k_n}\sum_{i=1}^{k_n}m(X_{(n,i)}(x)).
\end{equation}
Write the noise and approximation errors as
\begin{equation}
\label{noise-bias}
I_{n,2}=\EXP\int|m_n-\overline m_n|^2\,d\mu,
\qquad
J_{n,2}=\EXP\int|\overline m_n-m|^2\,d\mu.
\end{equation}
For fixed $n$ all these quantities are finite: Jensen's inequality
and the fact that the selected responses form a subset of the
sample give
\[
\EXP|m_n(X)|^2
\le\EXP\left[\frac1{k_n}\sum_{i=1}^{k_n}
|Y_{(n,i)}(X)|^2\right]
\le\frac n{k_n}\EXP Y^2.
\]
Conditional Jensen's inequality gives the same bound for $\overline m_n(X)$,
and $\EXP|m(X)|^2\le\EXP Y^2$.

Conditionally on the feature sample and query,
$m_n(X)-\overline m_n(X)$ has mean zero, whereas
$\overline m_n(X)-m(X)$ is fixed. Thus the cross term vanishes
when the squared error is expanded, giving
\begin{equation}
\label{quadratic-decomp}
R_{n,2}=I_{n,2}+J_{n,2}.
\end{equation}
The selected responses are conditionally independent. Their
variances therefore add:
\[
\EXP[(m_n(X)-\overline m_n(X))^2\mid X,X_1,\ldots,X_n]
=\frac1{k_n^2}\sum_{i=1}^{k_n}
\sigma^2(X_{(n,i)}(X)).
\]
Averaging over the feature sample and query yields the exact identity
(compare Lemma~\ref{averaging-lemma} above)
\begin{equation}
\label{noise-identity}
I_{n,2}=\frac1{k_n}A_n(\sigma^2)
=\frac1{k_n}\int\sigma^2f_n\,d\mu.
\end{equation}
Thus the density describes which conditional variances are sampled,
while $1/k_n$ describes the averaging of independent response noise.
For the approximation error, the same functional averages squared
errors in approximating $m$ at the selected neighbors. This leads
to the following criterion.

\begin{theorem}[Mean squared consistency]
\label{quadratic-criterion}
Assume \eqref{nkn} and \eqref{defhx}. Fix
$1<p\le q\le\infty$ and suppose $F_p<\infty$.
Then $D_{q,2}<\infty$ is necessary and sufficient for
$R_{n,2}\to0$ for every joint distribution with feature marginal
$\mu$ satisfying
\begin{equation}
\label{quadratic-moment-class}
m\in L_{2p'}(\mu),\qquad \sigma^2\in L_{q'}(\mu).
\end{equation}
If $D_{q,2}=\infty$, there is a single fixed law satisfying
\eqref{quadratic-moment-class}, with $m=0$, for which
$R_{n,2}\to\infty$ along a subsequence. This last assertion
does not require $F_p<\infty$.
\end{theorem}
The proof is given in Section~\ref{sec:conditional-moments}.

The approximation hypothesis and the noise condition have distinct
roles. Approximating $m$ by a bounded Lipschitz function $v$ produces
$A_n((m-v)^2)$, which is bounded by
$F_p\|m-v\|_{2p'}^2$. For the noise, splitting the conditional
variance at a level $L$ in \eqref{noise-identity} gives
\[
I_{n,2}\le\frac L{k_n}
+D_{q,2}\|\sigma^2\IND_{\{\sigma^2>L\}}\|_{q'}.
\]
The first term vanishes as $n\to\infty$ and the tail norm vanishes
as $L\to\infty$. A larger $q$ permits a weaker variance moment,
while requiring a stronger density norm. The proof of necessity
uses the same exact variance identity.

\begin{corollary}[Conditional mean and response moments]
\label{mean-response-moments}
Assume \eqref{nkn} and \eqref{defhx}. If
$1<p\le q\le\infty$, $F_p<\infty$, and $D_{q,2}<\infty$,
then $R_{n,2}\to0$ for every joint distribution with feature
marginal $\mu$ satisfying
\[
m\in L_{2p'}(\mu),\qquad \EXP|Y|^{2q'}<\infty.
\]
In particular, $F_p<\infty$ suffices whenever
$\EXP|Y|^{2p'}<\infty$.
\end{corollary}

\begin{proof}
Since $\sigma^2(X)\le\EXP[Y^2\mid X]$, conditional Jensen's inequality gives
$\int(\sigma^2)^{q'}\,d\mu\le\EXP|Y|^{2q'}$.
For the last assertion, take $q=p$ and use
$D_{p,2}\le F_p$ and
$\int|m|^{2p'}\,d\mu\le\EXP|Y|^{2p'}$.
\end{proof}

At $p=q=\infty$, this recovers the finite-second-moment condition
under Stone's domination. Finite exponents allow more concentration
of the selected-neighbor distribution at the cost of stronger conditions on the
response moments. The following consequences emphasize the fourth
moment and the usual regression model with additive noise.
\subsection{Concrete moment conditions}

Theorem~\ref{quadratic-criterion} gives the following particularly simple
consequence under a fourth response moment.

\begin{corollary}[A finite fourth response moment]
\label{cor:quadratic-fourth-moment}
Assume \eqref{nkn} and \eqref{defhx}. If $F_2<\infty$, then
\[
R_{n,2}\longrightarrow0
\]
for every joint distribution of $(X,Y)$ with feature marginal
$\mu$ and $\EXP Y^4<\infty$.
\end{corollary}

\begin{proof}
Conditional Jensen's inequality gives
\[
\int |m|^4\,d\mu\le\EXP Y^4,
\qquad
\int(\sigma^2)^2\,d\mu
\le\int\bigl(\EXP[Y^2\mid X=x]\bigr)^2\,\mu(dx)
\le\EXP Y^4.
\]
Thus $m\in L_4(\mu)$ and $\sigma^2\in L_2(\mu)$.
Moreover,
$\sup_n\|f_n\|_2/k_n\le F_2<\infty$.
The quadratic criterion applies with $p=q=2$.
\end{proof}

The fourth moment is one convenient way to control both the
regression function and the noise. The separate conditions in
the quadratic criterion also cover responses with only a
finite second moment.

\begin{corollary}[A fourth moment of the conditional mean]
\label{cor:quadratic-mixed-moments}
Assume \eqref{nkn} and \eqref{defhx}, and suppose $F_2<\infty$.
If $m\in L_4(\mu)$ and $\EXP Y^2<\infty$, then $R_{n,2}\to0$
under either of the following additional conditions:
\begin{enumerate}
\item
The conditional variance is essentially bounded.
\item
$\sup_n\|f_n\|_\infty/k_n<\infty$.
In particular, this holds if $\sup_n n/k_n^2<\infty$.
\end{enumerate}
\end{corollary}

\begin{proof}
The approximation term tends to zero because
$F_2<\infty$ and $m\in L_4(\mu)$.
If $\sigma^2\le K$, then its noise term satisfies
\[
I_{n,2}=\frac1{k_n}\int\sigma^2 f_n\,d\mu
\le\frac K{k_n}\longrightarrow0.
\]
For the second assertion, put
$D=\sup_n\|f_n\|_\infty/k_n$.
Since $\sigma^2\in L_1(\mu)$, for every $L>0$,
\[
I_{n,2}
\le\frac L{k_n}
+D\int\sigma^2\IND_{\{\sigma^2>L\}}\,d\mu.
\]
First let $n\to\infty$, and then let $L\to\infty$.
The bound $f_n\le n/k_n$ proves the last assertion.
\end{proof}

For example, the first case applies to additive noise
$Y=m(X)+\varepsilon$, where $\varepsilon$ is independent of
$X$, has mean zero and finite variance. No fourth moment of
$\varepsilon$ is required. The second case allows unbounded
conditional variances while retaining only a finite second
response moment; it controls their contribution through the
scaled density bound.

\subsection{Neighborhoods of fixed population mass}
\label{sec:fixed-mass-geometry}
A condition on $\mu$ can also be formulated through
neighborhoods of fixed population mass. For $0<u\le1$, define
\begin{equation}
\label{fixed-mass-reverse}
f_u^*(z)=\frac1u\mu\{x:H_x(\rho(x,z))\le u\}.
\end{equation}
Thus $f_u^*(z)$ is the mass, divided by $u$, of query centres
whose ball reaches $z$ before accumulating more than $u$
feature probability. Under \eqref{defhx},
$H_x(\rho(x,Z))$ is uniform on $[0,1]$ when $Z\sim\mu$.
Tonelli's theorem therefore gives $\int f_u^*\,d\mu=1$.

There is an exact relation with the sample densities. Let
$V_n$ be an auxiliary random variable with the beta distribution
with parameters $k_n+1$ and $n-k_n$. Then
\begin{equation}
\label{density-beta-mixture}
f_n(z)=\EXP[f_{V_n}^*(z)],
\end{equation}
where the expectation is only over $V_n$. Indeed, if
$U_n\sim\operatorname{Beta}(k_n,n-k_n)$, the binomial
selection probability in \eqref{gn} equals
$\PROB\{U_n\ge H_x(\rho(x,z))\}$.
Substitution into \eqref{fn} and Tonelli's theorem give
\[
f_n(z)=\frac n{k_n}\EXP[U_n f_{U_n}^*(z)].
\]
To identify the weighted distribution explicitly, let
$\mathrm B(a,b)=\int_0^1 u^{a-1}(1-u)^{b-1}\,du$ be the
beta function. The density of $U_n$ is
\[
b_{U_n}(u)=
\frac{u^{k_n-1}(1-u)^{n-k_n-1}}
{\mathrm B(k_n,n-k_n)},\qquad 0<u<1;
\]
see \citet[Chapter~1]{BiDe15}.
Since $\mathrm B(k_n+1,n-k_n)=(k_n/n)\mathrm B(k_n,n-k_n)$,
\[
\frac n{k_n}\,u\,b_{U_n}(u)
=\frac{u^{k_n}(1-u)^{n-k_n-1}}
{\mathrm B(k_n+1,n-k_n)}
=b_{V_n}(u).
\]
The right side is the probability density of
$\operatorname{Beta}(k_n+1,n-k_n)$. Consequently,
\[
f_n(z)
=\frac n{k_n}\EXP[U_n f_{U_n}^*(z)]
=\int_0^1 f_u^*(z)b_{V_n}(u)\,du
=\EXP[f_{V_n}^*(z)],
\]
which proves \eqref{density-beta-mixture}. Thus $f_n$
averages the fixed-mass densities over these auxiliary mass
thresholds.

\begin{corollary}[A fixed-mass condition]
\label{cor:fixed-mass-fourth-moment}
Assume \eqref{nkn} and \eqref{defhx}, and suppose that
\begin{equation}
\label{fixed-mass-L2-sufficient}
\limsup_{u\downarrow0}\int\bigl(f_u^*(z)\bigr)^2\,\mu(dz)
<\infty.
\end{equation}
Then $R_{n,2}\to0$ for every joint distribution with feature
marginal $\mu$, $\EXP Y^2<\infty$, and $m\in L_4(\mu)$
satisfying
\begin{equation}
\label{fixed-mass-variance}
\limsup_{u\downarrow0}\int\sigma^2(z)f_u^*(z)\,\mu(dz)
<\infty.
\end{equation}
In particular, \eqref{fixed-mass-L2-sufficient} implies
$R_{n,2}\to0$ for every joint distribution with feature
marginal $\mu$ and $\EXP Y^4<\infty$.
\end{corollary}

\begin{proof}
There is $u_0\in(0,1)$ such that both integrals in the
hypotheses are bounded uniformly for $0<u<u_0$.
For $u_0\le u\le1$, the identities $0\le f_u^*\le1/u$ and
$\int f_u^*\,d\mu=1$ give
\[
\|f_u^*\|_2^2\le\frac1u\le\frac1{u_0},
\qquad
\int\sigma^2 f_u^*\,d\mu
\le\frac1{u_0}\int\sigma^2\,d\mu<\infty.
\]
The square-integral hypothesis and its bound away from zero give
\[
M_2=\sup_{0<u\le1}\|f_u^*\|_2<\infty.
\]
Likewise, the variance hypothesis and $\EXP Y^2<\infty$ give
\[
M_\sigma=\sup_{0<u\le1}\int\sigma^2 f_u^*\,d\mu<\infty.
\]
The beta-mixture identity \eqref{density-beta-mixture},
Jensen's inequality, and Tonelli's theorem imply
\[
\|f_n\|_2^2
\le\EXP\|f_{V_n}^*\|_2^2\le M_2^2,
\qquad
A_n(\sigma^2)
=\EXP\int\sigma^2 f_{V_n}^*\,d\mu\le M_\sigma.
\]
Thus $F_2\le M_2$, so the approximation argument in
Theorem~\ref{quadratic-criterion} gives $J_{n,2}\to0$
for $m\in L_4(\mu)$. The exact noise identity gives
\[
I_{n,2}=\frac1{k_n}A_n(\sigma^2)
\le\frac{M_\sigma}{k_n}\longrightarrow0.
\]
Together these prove the first assertion.
If $\EXP Y^4<\infty$, conditional Jensen's inequality gives
$m\in L_4(\mu)$ and $\sigma^2\in L_2(\mu)$.
Cauchy--Schwarz then yields
\[
\int\sigma^2 f_u^*\,d\mu
\le\|\sigma^2\|_2\|f_u^*\|_2\le\|\sigma^2\|_2M_2,
\]
so the variance condition also holds.
\end{proof}

\begin{lemma}[Comparison at deterministic mass levels]
\label{deterministic-mass-comparison}
Assume \eqref{defhx}, and put $c_0=1-\log2>0$.
For every $n$ such that $2k_n<n$ and every $z\in\X$,
\begin{equation}
\label{deterministic-mass-bounds}
\frac14 f^*_{k_n/(2n)}(z)
\le f_n(z)
\le 2f^*_{2k_n/n}(z)
+\frac{n}{2k_n}e^{-c_0k_n}.
\end{equation}
The lower bound holds for every $n\ge2$.
\end{lemma}

\begin{proof}
Put $u=k_n/(2n)$. If $H_x(\rho(x,z))\le u$, the number
of other observations closer to $x$ than $z$ has expectation
at most $(n-1)u$. Markov's inequality gives
$g_n(x,z)\ge1-(n-1)u/k_n\ge1/2$. Integrating over these
query centres yields
\[
f_n(z)\ge\frac{n}{2k_n}\mu\{x:H_x(\rho(x,z))\le u\}
=\frac14 f_u^*(z).
\]

For the upper bound, put $v=2k_n/n<1$ and use
\eqref{density-beta-mixture}. By definition, $u f_u^*(z)$
is nondecreasing in $u$, so $f_u^*(z)\le(v/u)f_v^*(z)$
for $u\le v$. Since $\EXP[V_n^{-1}]=n/k_n$,
\[
\EXP[f_{V_n}^*(z)\IND_{\{V_n\le v\}}]
\le v f_v^*(z)\EXP[V_n^{-1}]=2f_v^*(z).
\]
The variable $V_n$ has the distribution of the $(k_n+1)$-st
order statistic of $n$ independent uniforms. Thus, for
$B\sim\operatorname{Bin}(n,v)$,
\begin{align*}
\PROB(V_n>v)
&=\PROB(B\le k_n)
\le2^{k_n}\EXP[2^{-B}]\\
&=2^{k_n}(1-v/2)^n
\le\exp(k_n\log2-nv/2)=e^{-c_0k_n}.
\end{align*}
On $\{V_n>v\}$, $f_{V_n}^*(z)\le1/v$, so the remaining
part of the mixture is at most
\[
\EXP[f_{V_n}^*(z)\IND_{\{V_n>v\}}]
\le\frac1v e^{-c_0k_n}
=\frac{n}{2k_n}e^{-c_0k_n}.
\]
Adding the two parts proves the upper bound.
\end{proof}

In Euclidean space, the fixed-mass densities are bounded by the
same cone geometry that underlies Stone's lemma
\citep[Lemma~6.2]{GyKoKrWa02}. Stone's argument
bounds how often a sample point can be selected as a neighbor;
the cones used in the proof below bound the $\mu$-mass of query centres whose balls
reach a fixed point with mass at most $u$. The resulting uniform
bound on $f_u^*$ transfers to $f_n$ through
\eqref{density-beta-mixture}, recovering Stone's domination
condition \eqref{stone}.

\begin{lemma}[Euclidean fixed-mass bound]
\label{lem:euclidean-fixed-mass}
Let $\X=\mathbb R^d$ with Euclidean distance and assume
\eqref{defhx}. There is a constant $\gamma_d<\infty$,
depending only on $d$, such that
\[
f_u^*(z)\le\gamma_d,
\qquad z\in\mathbb R^d,\quad 0<u\le1.
\]
\end{lemma}

This is the reverse-ball estimate of
Devroye and Gy\"orfi~\cite[Lemma~1, p.~260]{DeGy85}.
We include its geometric argument.

\begin{proof}
Cover $\mathbb R^d\setminus\{z\}$ by $\gamma_d$ punctured cones
with vertex $z$, chosen so that the angle between any two
rays in one cone is strictly less than $\pi/3$.
Such a finite cover exists by compactness of the unit sphere.
If $x$ and $y$ lie in one cone and
$0<\|y-z\|\le\|x-z\|$, then
\[
\|x-y\|<\|x-z\|.
\]
For one cone $C$, let
\[
E_C=\{x\in C:\mu(B(x,\|x-z\|))\le u\}.
\]
For every $x\in E_C$, its radial initial segment satisfies
\[
E_C\cap B(z,\|x-z\|)\subseteq B(x,\|x-z\|),
\]
and therefore has $\mu$-mass at most $u$.
If $E_C$ is empty, its mass is zero. Otherwise, choose a
sequence of radii of points of $E_C$ increasing
cofinally to the supremum of these radii, including the
maximum if it is attained. The associated initial segments
increase to $E_C$, so $\mu(E_C)\le u$.
Summing over the cones and using $\mu\{z\}=0$ gives
\[
\mu\{x:H_x(\|x-z\|)\le u\}\le\gamma_d u.
\]
Divide by $u$.
\end{proof}

For functional predictors, an important question is to verify these
density conditions from explicit properties of the predictor law.
For example, when continuous curves are viewed as elements of
$L_2([0,1])$, which distributions satisfy
\eqref{fixed-mass-L2-sufficient}? The criterion gives a sufficient
distributional condition for every fourth-moment response, without
requiring the conditional mean to be smooth.

\subsection{Noiseless observations and a fixed number of neighbors}
A fixed number of neighbors need not average away response noise.
For example, if $Y$ is an independent Rademacher sign, then
$m=0$ and the mean squared error is exactly $1/k$ when $k$
neighbors are averaged. For noiseless observations, however,
the noise term vanishes, and consistency only requires the
neighbor locations to approach the query while controlling
approximation error.

\begin{corollary}[Noiseless observations]
\label{cor:noiseless-fixed-neighbors}
Assume \eqref{defhx} and \eqref{fixed-mass-L2-sufficient}.
Let $1\le k_n<n$ satisfy $k_n/n\to0$; it is not necessary
that $k_n\to\infty$. Then for every noiseless response
$Y=m(X)$ with $m\in L_4(\mu)$,
\[
R_{n,2}\longrightarrow0.
\]
In particular, this conclusion holds for $1$-NN and for every
other fixed number of neighbors, once $n$ exceeds that number.
\end{corollary}

\begin{proof}
Condition \eqref{fixed-mass-L2-sufficient} and the bound
$\|f_u^*\|_2^2\le1/u$ give
$M=\sup_{0<u\le1}\|f_u^*\|_2<\infty$.
For every $1\le k_n<n$, the beta-mixture identity
\eqref{density-beta-mixture} and Jensen's inequality yield
\[
\|f_n\|_2^2
\le\EXP\|f_{V_n}^*\|_2^2\le M^2.
\]
Neither step requires $k_n\to\infty$. Thus $F_2\le M$.

We also verify locality under the weaker assumption on $k_n$.
For $\mu$-almost every $x$, every positive-radius ball around
$x$ has positive $\mu$-probability, by separability of the
feature space. Fix such an $x$ and $b>0$, and put
$p_x=\mu(B(x,b/2))>0$. The number $N_n(x)$ of sample
points in this ball has binomial distribution with parameters
$n,p_x$. Since $k_n/n\to0$, the law of large numbers gives
\[
\PROB\{\rho(x,X_{(n,k_n)}(x))\ge b\}
\le\PROB\{N_n(x)<k_n\}\longrightarrow0.
\]
Dominated convergence over the independent query gives the
same convergence with $x$ replaced by $X$.

Fix $\varepsilon>0$ and choose a bounded Lipschitz function
$v$ with $\|m-v\|_4<\varepsilon$. Such an approximation is
provided by Lemma~\ref{approximation-locality} below. Put $\phi=m-v$.
Since the response is noiseless, $m_n=\overline m_n$, so
adding and subtracting the neighbor average of $v$ gives
\begin{align*}
R_{n,2}\le3\biggl[&A_n(\phi^2)+\|\phi\|_2^2\\
&+\EXP\int\left|\frac1{k_n}\sum_{i=1}^{k_n}
v(X_{(n,i)}(x))-v(x)\right|^2\mu(dx)\biggr].
\end{align*}
The first two terms satisfy
\[
A_n(\phi^2)\le F_2\|\phi\|_4^2<M\varepsilon^2,
\qquad \|\phi\|_2^2\le\|\phi\|_4^2<\varepsilon^2.
\]
For every $\eta>0$, uniform continuity supplies $b>0$ such
that $|v(z)-v(x)|\le\eta$ whenever $\rho(x,z)<b$.
Consequently, the remaining term is at most
\[
\eta^2+4\|v\|_\infty^2
\PROB\{\rho(X,X_{(n,k_n)}(X))\ge b\}.
\]
Locality, followed by $\eta\downarrow0$, makes this term tend
to zero. Hence
$\limsup_n R_{n,2}\le3(M+1)\varepsilon^2$.
Letting $\varepsilon\downarrow0$ proves the assertion.
\end{proof}

The proof only uses the fixed-mass assumption to obtain
$F_2<\infty$. A uniform bound $F_\infty<\infty$ instead
permits the same proof with $m\in L_2(\mu)$, since
$A_n(\phi^2)\le F_\infty\|\phi\|_2^2$ and approximation in
$L_2(\mu)$ suffices. This recovers the corresponding noiseless
Euclidean result, including fixed $k$, under the standing
no-distance-tie assumption; see \citet[Problem~6.3]{GyKoKrWa02}.

\section{Nagata dimension and nearest-neighbor selection}
\label{sec:nagata}
The density criteria separate the distributional requirement from
its geometric verification. Global geometric control yields Stone's
domination. Sigma-finite dimensionality does not ensure consistency
for unbounded responses, as the example below shows.

\subsection{Global dimension, scale, and relative dimension}

Recall that the \emph{multiplicity} of a family $\gamma$ of
subsets of $\X$ is
\[
\sup_{z\in\X}\sum_{V\in\gamma}\IND_V(z).
\]
Also, a family of balls is \emph{disconnected} if no ball of
the family contains the centre of any other ball.

\begin{definition}[Nagata dimension]
\label{def:nagata}
Let $\mathcal A\subseteq\X$, $s\in(0,\infty]$, and
$\delta\in\mathbb N_0$. The subspace $\mathcal A$ has Nagata
dimension at most $\delta$ in $\X$ on scale $s$ if every finite
family of closed balls in $\X$ with centres in $\mathcal A$
and radii less than $s$ admits a subfamily of multiplicity at
most $\delta+1$ in $\X$ which covers all the centres of the
original balls. Write $\dim_{\mathrm{Nag}}^s(\mathcal A,\X)$
for the smallest such $\delta$, or $\infty$ if none exists.
\end{definition}

An equivalent condition is that every finite disconnected
family of such balls has multiplicity at most $\delta+1$.
Closed balls can also be replaced with open balls, or a mixture
of the two, cf. \citet[Propositions~7.2--7.3]{CoKuPe20}.
For relative dimension the multiplicity is measured throughout
the \emph{ambient} space $\X$, not only on $\mathcal A$.
An intrinsic dimension bound on $\mathcal A$ therefore does
not by itself establish the required relative bound.

We distinguish the original, global notion
\[
\dim_{\mathrm{Nag}}(\X)
:=\dim_{\mathrm{Nag}}^\infty(\X,\X)
\]
from finite dimension on some positive scale. The latter
restricts the covering property to small balls; the global
notion imposes one common bound for balls of every radius.
The relative, scale-dependent formulation is due to Preiss
\cite{preiss83}. For a normed vector space, homotheties preserve
centre containment and multiplicity while scaling radii, so
a bound on one positive scale gives the same global bound.

Following Preiss, a metric space is \emph{sigma-finite
dimensional in the sense of Nagata} if it is a countable union
of subspaces $\mathcal A_j$, each having finite Nagata dimension
in $\X$ on some scale $s_j>0$. The dimensions and scales may
all be different. The subspaces may be taken closed, hence
Borel \cite[Proposition~7.4, Definition~7.6 and Remark~7.7]{CoKuPe20}.
For basic properties and further examples see
\citet[Section~3]{Pes26}.

\subsection{Global domination}
In a complete separable space with global Nagata dimension at
most $\delta$, the covering argument gives
$f_u^*\le\delta+1$ for every $u$, and
\eqref{density-beta-mixture} gives $f_n\le\delta+1$.
The fixed-mass covering argument of
\citet[Lemma~4.6]{KuPe24} is used here at the infinite scale;
its proof in this case is given in
Section~\ref{sec:nagata-proofs}. For a finite scale $s$, the
covering bound applies only to centres with $\rho(x,z)<s$.
Small ball mass alone does not enforce this radius restriction
uniformly over the centres. This distinction is essential to
the counterexample below.

\begin{corollary}[Global Nagata dimension]
\label{nagata-global-consistency}
Assume \eqref{nkn} and \eqref{defhx}. If
$\dim_{\mathrm{Nag}}(\X)\le\delta<\infty$, then
\[
F_\infty\le\delta+1.
\]
Thus Stone's condition \eqref{stone} holds with $C=\delta+1$,
and $k_n$-NN is weakly consistent in $L_2(\mu)$ for every joint distribution
of $(X,Y)$ with feature marginal $\mu$ and $\EXP Y^2<\infty$.
\end{corollary}

\subsection{Unbounded responses on the porcupine}

\begin{proposition}
\label{nagata-unbounded-counterexample}
There is a complete separable metric probability space
$(\mathcal T,\rho,\mu)$ that is sigma-finite dimensional in the
sense of Nagata and satisfies \eqref{defhx}, with the following
property. There is one fixed nonnegative noiseless response
$Y=m(X)$ with $\EXP|Y|^r<\infty$ for every finite $r\ge1$ such that
\[
R_{n,r}:=\EXP\int|m_n-m|^r\,d\mu\longrightarrow\infty
\]
for every finite $r\ge1$ and every sequence $1\le k_n<n$
satisfying $k_n/n\to0$. The joint distribution of $(X,Y)$
depends neither on
$r$ nor on the neighbor sequence.
\end{proposition}

The space $\mathcal T$ is the porcupine, or hedgehog, formed by
countably many copies of $[0,1]$, called spines, whose zero
endpoints are identified as a common root $o$. Write $(i,a)$
for the point at distance $a\in(0,1]$ from $o$ on spine $i\ge2$.
Two points have distance $|a-b|$ on the same spine and $a+b$
on different spines. Each spine has Nagata dimension one
relative to $\mathcal T$, although $\mathcal T$ itself has
infinite Nagata dimension at every positive scale. For the explicit
unbounded response in Proposition~\ref{nagata-unbounded-counterexample},
$k$-NN regression is inconsistent in every finite moment order.
The regression function is
$m(o)=0$ and $m((i,a))=[\log(e/a)]^3$.

In this same construction, $f_n$ becomes large near the root
on regions of vanishing feature probability. The regression
function $m$ specified above is large enough on these regions
to make the expected integrated error diverge.

In a finite-dimensional normed space, Stone's geometric bound
gives $F_\infty<\infty$ for every feature marginal $\mu$
satisfying \eqref{defhx}. Taking $p=q=\infty$ in
Corollary~\ref{mean-response-moments} therefore gives weak
$L_2(\mu)$ consistency whenever $\EXP Y^2<\infty$. The same conclusion holds under the
global Nagata hypothesis of
Corollary~\ref{nagata-global-consistency}, which imposes one
common dimension bound on balls of every radius. Thus the
failure for unbounded responses above cannot occur under such
a global density bound.

By contrast, imposing the dimension bound only on balls of
radius less than one fixed $s>0$ does not by itself ensure
mean squared consistency, even for a response with all finite
moments. The separated-spine construction in
Remark~\ref{finite-scale-unbounded} has Nagata dimension at most
one on scale $1/2$, yet admits a fixed noiseless response with
all finite moments for which $R_{n,r}\to\infty$ for every
finite $r\ge1$.

Neither sigma-finite Nagata dimension nor finite Nagata dimension
on one positive scale guarantees the uniform $L_p(\mu)$ bounds
assumed in the sufficient conditions of Section~\ref{sec:consistency}.
In each of the two examples, for every sequence satisfying
\eqref{nkn},
\[
F_p=\sup_{n\ge2}\|f_n\|_p=\infty,\qquad 1<p\le\infty.
\]
Indeed, the proofs show that $\int m f_n\,d\mu\to\infty$,
whereas $m\in L_{p'}(\mu)$. H\"older's inequality gives
\[
\int m f_n\,d\mu\le\|m\|_{p'}\|f_n\|_p,
\]
so $\|f_n\|_p$ cannot remain bounded as $n$ increases.

The proofs of the geometric statements are given in
Section~\ref{sec:nagata-proofs}.

\section{Smoothness and convergence rates}
\label{sec:rates}
The preceding criteria impose conditions on neighbor selection and
allow arbitrary conditional means in the stated moment class.
A complementary route imposes smoothness directly on $m$.
The next bound requires no density-norm condition.

\subsection{A generalized Lipschitz condition}

Quantitative bounds require regularity of the regression
function. One useful condition measures smoothness through
the probability mass of a neighborhood rather than its
radius. This results in a joint smoothness and tail condition. Following \citet{GyWe21}, suppose that a nondecreasing
function $h:[0,1]\to[0,\infty)$ with $h(0)=0$ and
$h(u)\to0$ as $u\downarrow0$ satisfies
\begin{equation}
\label{gen-Lip}
|m(x)-m(z)|\le h\bigl(H_x(\rho(x,z))\bigr),
\qquad x,z\in\X.
\end{equation}

Mass-based smoothness conditions have been used to obtain
classification rates without requiring bounded feature support.
The modified Lipschitz condition in \citet{DoGyWa18} compares
regression values at two points using the probability of the ball
joining them. \citet{ChDa14} uses a related condition comparing a
regression value with its average over a ball. The generalized Lipschitz 
condition \eqref{gen-Lip} follows the metric-space formulation
in \citet{GyWe21}.

\begin{proposition}[A mean squared error bound]
\label{regr_rate}
Assume \eqref{defhx}, \eqref{gen-Lip}, and
$\sigma^2\le K$ almost everywhere. If $g=h^2$ is concave,
then, for every $1\le k_n<n$,
\begin{equation}
\label{quadratic-rate}
R_{n,2}\le\frac K{k_n}
+h\left(\frac{k_n}{n+1}\right)^2
\le\frac K{k_n}+h\left(\frac{k_n}{n}\right)^2.
\end{equation}
\end{proposition}

\begin{proof}
The exact quadratic decomposition gives
$R_{n,2}=I_{n,2}+J_{n,2}$, with $I_{n,2}\le K/k_n$.
Fix a query $x$. The probability integral transform and
\eqref{defhx} imply that
$H_x(\rho(x,X_i))$, $1\le i\le n$, are independent
uniform random variables. Consequently,
\[
U_{(n,k_n)}(x)
=H_x\bigl(\rho(x,X_{(n,k_n)}(x))\bigr)
\]
has the $k_n$-th uniform order-statistic distribution, whose
mean is $k_n/(n+1)$. By \eqref{gen-Lip}, monotonicity of $h$,
and Jensen's inequality for the neighbor average,
\begin{align*}
\left|\frac1{k_n}\sum_{i=1}^{k_n}
m(X_{(n,i)}(x))-m(x)\right|^2
&\le\frac1{k_n}\sum_{i=1}^{k_n}
|m(X_{(n,i)}(x))-m(x)|^2\\
&\le g\bigl(U_{(n,k_n)}(x)\bigr).
\end{align*}
Taking expectations and applying concavity gives
\[
J_{n,2}\le\int\EXP g\bigl(U_{(n,k_n)}(x)\bigr)\,\mu(dx)
\le g\left(\frac{k_n}{n+1}\right).
\]
This proves the first inequality; the second uses monotonicity.
\end{proof}

For example, if $h(u)=c_0u^a$ with $0<a\le1/2$, then
\[
R_{n,2}\le\frac K{k_n}
+c_0^2\left(\frac{k_n}{n+1}\right)^{2a}.
\]
Choosing $k_n$ of order $n^{2a/(2a+1)}$ gives the rate
$R_{n,2}=O(n^{-2a/(2a+1)})$.
In particular, if $\X$ is the $d$-dimensional Euclidean space and $a=1/d$ with $d\ge2$, then the order is $O(n^{-2/(d+2)})$ under the stated mass-based smoothness
condition. Euclidean geometry alone does not imply that
smoothness condition.

\section{Proofs of the consistency and geometric results}
\label{sec:proofs}

\subsection{The averaging and selection-count identities}
\label{sec:averaging-proofs}

\begin{proof}[Proof of Lemma~\ref{averaging-lemma}]
Fix a nonnegative Borel function $\varphi$. We first work with a
fixed query $x$. For every realization of the sample, summing
over the selected neighbors is the same as summing over all
sample indices and retaining only those selected:
\[
\frac1{k_n}\sum_{i=1}^{k_n}\varphi(X_{(n,i)}(x))
=
\frac1{k_n}\sum_{j=1}^n
\varphi(X_j)\IND_{\{X_j\text{ is selected at }x\}}.
\]

We next compute the expectation of each summand on the right.
Fix an index $j$ and condition on $X_j=z$. By independence, the
remaining $n-1$ observations still have the joint law
$\mu^{n-1}$. For this fixed $x$ and $z$, the no-ties assumption
implies that none of the remaining observations has distance
$\rho(x,z)$ from $x$, almost surely. Thus the candidate $z$ is
selected precisely when at most $k_n-1$ of the remaining
observations are closer to $x$.

This conditional selection probability is $g_n(x,z)$.
Although \eqref{gn} uses the indices $2,\ldots,n$, the remaining
observations have the same joint law after removing any index
$j$.
Consequently, for $\mu$-almost every $z$,
\[
\EXP\bigl[
\varphi(X_j)\IND_{\{X_j\text{ is selected at }x\}}
\mid X_j=z
\bigr]
=\varphi(z)g_n(x,z).
\]
Here the location $z$ is held fixed, and the expectation averages
only over the remaining observations. Averaging this conditional
expectation over $X_j$, whose distribution is $\mu$, gives
\[
\EXP\bigl[
\varphi(X_j)\IND_{\{X_j\text{ is selected at }x\}}
\bigr]
=
\int\varphi(z)g_n(x,z)\,\mu(dz).
\]
The right-hand side does not depend on $j$. Taking expectations
in the first identity and summing these $n$ identical integrals
therefore gives
\[
\frac1{k_n}\sum_{i=1}^{k_n}\EXP\varphi(X_{(n,i)}(x))
=
\frac n{k_n}\int\varphi(z)g_n(x,z)\,\mu(dz).
\]
This explains both the use of $g_n$ and the factor $n/k_n$.

Finally, the independent query has distribution $\mu$.
Integrating over its location and applying Tonelli's theorem,
which is applicable because all integrands are nonnegative,
we obtain
\begin{align*}
A_n(\varphi)
&=
\int\left(
\frac n{k_n}\int\varphi(z)g_n(x,z)\,\mu(dz)
\right)\mu(dx)\\
&=
\int\varphi(z)
\left(\frac n{k_n}\int g_n(x,z)\,\mu(dx)\right)\mu(dz)\\
&=
\int\varphi(z)f_n(z)\,\mu(dz),
\end{align*}
where the last equality uses \eqref{fn}. These equalities remain
valid when the integrals are infinite.

Since $0\le g_n\le1$, the definition of $f_n$ gives
$0\le f_n\le n/k_n$. Taking $\varphi=1$ in the averaging
identity gives
\[
\int f_n\,d\mu=A_n(1)=1,
\]
because exactly $k_n$ observations are selected.

For a signed $\varphi\in L_1(\mu)$, the bound on $f_n$ gives
$\int|\varphi|f_n\,d\mu<\infty$. Applying the identity to the
positive and negative parts of $\varphi$ therefore proves the
signed version. In particular, for every fixed $n$, $A_n$ is a
continuous linear functional on $L_r(\mu)$ for every
$1\le r\le\infty$.
\end{proof}

\begin{lemma}[Selection-count identity]
\label{selection-count-identity}
Fix $n\ge2$, and let $Z_0,\ldots,Z_n$ be independent with
distribution $\mu$. For each $i$, let $\mathcal N_i$ be the
indices of the $k_n$ nearest neighbors of query $Z_i$ among
the other $n$ observations. Define the selection count of
observation $Z_j$ by
\begin{equation}
\label{augmented-selection-count}
D_{n,j}=\sum_{\substack{0\le i\le n\\i\ne j}}
\IND_{\{j\in\mathcal N_i\}}.
\end{equation}
Under \eqref{defhx}, for every $j\in\{0,\ldots,n\}$,
\begin{equation}
\label{density-selection-count}
f_n(z)=\frac1{k_n}\EXP[D_{n,j}\mid Z_j=z]
\qquad\text{for $\mu$-almost every $z$}.
\end{equation}
In particular, $\EXP D_{n,j}=k_n$.
\end{lemma}

\begin{proof}[Proof of Lemma~\ref{selection-count-identity}]
Fix $j$ and a possible query index $i\ne j$. Conditional on
$Z_i=x$ and $Z_j=z$, the remaining $n-1$ feature observations
are independent with distribution $\mu$. By \eqref{defhx},
none has distance $\rho(x,z)$ from $x$, almost surely.
Consequently,
\[
\PROB\{j\in\mathcal N_i\mid Z_i=x,Z_j=z\}=g_n(x,z).
\]
Independence of $Z_i$ and $Z_j$ gives
\[
\EXP[\IND_{\{j\in\mathcal N_i\}}\mid Z_j=z]
=\int g_n(x,z)\,\mu(dx)
\qquad\text{for $\mu$-almost every $z$}.
\]
There are $n$ possible query indices $i\ne j$. Summing their
contributions and using \eqref{fn},
\[
\EXP[D_{n,j}\mid Z_j=z]
=n\int g_n(x,z)\,\mu(dx)=k_n f_n(z).
\]
Integrating over $z$ and using $\int f_n\,d\mu=1$ gives
$\EXP D_{n,j}=k_n$.
\end{proof}

\subsection{Approximation and the quadratic criterion}
\label{sec:conditional-moments}
We first record the approximation fact used below.
\begin{lemma}
\label{approximation-locality}
Bounded Lipschitz functions are dense in $L_a(\mu)$ for every
$1\le a<\infty$.
\end{lemma}
\begin{proof}
For Lipschitz density in $L_1(\mu)$, see
\citet[supplementary material, Lemma~A.1]{HaKoSaWe21}.
We give a proof for all $1\le a<\infty$.
Let $\mathcal H$ be their closure in $L_a(\mu)$. For a nonempty
closed set $F$, the bounded Lipschitz functions
$\max\{0,1-j\rho(x,F)\}$ converge pointwise to $\IND_F$.
Dominated convergence shows that $\IND_F\in\mathcal H$.
The class of Borel sets whose indicators belong to $\mathcal H$
is a Dynkin class: it contains $\X$, is closed under complements,
and is closed under countable disjoint unions by $L_a$ convergence
of finite partial sums. The closed sets form a generating
$\pi$-system, so this class contains every Borel set. Density of
simple functions proves the assertion.
\end{proof}

\begin{proof}[Proof of Theorem~\ref{quadratic-criterion}]
First suppose $D_{q,2}<\infty$ and fix a joint law satisfying
\eqref{quadratic-moment-class}. The moment assumptions imply
\[
\EXP Y^2=\int\sigma^2\,d\mu+\int m^2\,d\mu
\le\|\sigma^2\|_{q'}+\|m\|_{2p'}^2<\infty,
\]
so the quadratic identities above apply.

For the approximation term, fix $\varepsilon>0$ and choose a
bounded Lipschitz function $v$ with
$\|m-v\|_{2p'}<\varepsilon$. This is possible because $p>1$
and therefore $2p'<\infty$. Put $\phi=m-v$. Adding and subtracting
the neighbor average of $v$ and its value at the query gives
\[
\overline m_n(x)-m(x)
=\frac1{k_n}\sum_{i=1}^{k_n}\phi(X_{(n,i)}(x))
+\left[\frac1{k_n}\sum_{i=1}^{k_n}
v(X_{(n,i)}(x))-v(x)\right]-\phi(x).
\]
The three terms are the approximation error at the selected
neighbors, the local averaging error for $v$, and the negative
approximation error at the query. The inequality for a squared
sum and Jensen's inequality for the first average yield
\begin{equation}
\label{bias-bound}
J_{n,2}\le3\left[
A_n(\phi^2)
+\EXP\int\left|\frac1{k_n}\sum_{i=1}^{k_n}
v(X_{(n,i)}(x))-v(x)\right|^2\mu(dx)
+\|\phi\|_2^2\right].
\end{equation}
The averaging identity (Lemma~\ref{averaging-lemma}) and H\"older's inequality give
\begin{equation}
\label{bias-holder}
A_n(\phi^2)=\int \phi^2f_n\,d\mu
\le F_p\|\phi\|_{2p'}^2<F_p\varepsilon^2.
\end{equation}
Since $\mu(\X)=1$, H\"older's inequality also gives
\[
\|\phi\|_2^2=\int \phi^2\cdot1\,d\mu
\le\|\phi^2\|_{p'}\|1\|_p
=\|\phi\|_{2p'}^2<\varepsilon^2.
\]
For the middle term, let
$d_n(x)=\rho(x,X_{(n,k_n)}(x))$. For every $\eta>0$, bounded
uniform continuity of $v$ supplies $b>0$ such that
$|v(z)-v(x)|\le\eta$ whenever $\rho(x,z)<b$. Therefore
\begin{equation}
\label{regular-locality}
\EXP\int\left|\frac1{k_n}\sum_{i=1}^{k_n}
v(X_{(n,i)}(x))-v(x)\right|^2\mu(dx)
\le\eta^2+4\|v\|_\infty^2\PROB\{d_n(X)\ge b\}.
\end{equation}
Locality \eqref{locality} and dominated convergence make the
probability tend to zero; see also the locality argument of
\citet[Section~V]{CoHa67}. Letting $\eta\downarrow0$ and then
$\varepsilon\downarrow0$ in \eqref{bias-bound} proves
$J_{n,2}\to0$.

For the noise, the exact identity \eqref{noise-identity} gives
\begin{equation}
\label{quadratic-noise-truncation}
I_{n,2}\le\frac L{k_n}
+D_{q,2}\|\sigma^2\IND_{\{\sigma^2>L\}}\|_{q'}.
\end{equation}
The bounded part uses $A_n(1)=1$, and the tail uses
H\"older's inequality. Since $q'<\infty$, the tail norm tends
to zero as $L\to\infty$. First letting $n\to\infty$ at fixed
$L$, then letting $L\to\infty$, proves $I_{n,2}\to0$.
Together with \eqref{quadratic-decomp}, this proves sufficiency.

For necessity, define positive continuous linear functionals on
$L_{q'}(\mu)$ by
\[
\Lambda_n(\varphi)=\frac1{k_n}\int \varphi f_n\,d\mu.
\]
Their norms are $\|\Lambda_n\|=\|f_n\|_q/k_n$. Each is finite,
since $f_n/k_n\le n/k_n^2$.
For a nonnegative $\varphi\in L_{q'}(\mu)$, let $S$ be an independent
Rademacher sign and set $Y=S\sqrt{\varphi(X)}$.
This law has $m=0$ and $\sigma^2=\varphi$, and hence belongs to
\eqref{quadratic-moment-class}. For this law,
\[
R_{n,2}=I_{n,2}=\Lambda_n(\varphi).
\]
Consistency for every such law implies that $(\Lambda_n(\varphi))_n$
is bounded for every nonnegative $\varphi\in L_{q'}(\mu)$.
Decomposing a signed function into its positive and negative
parts, the uniform boundedness principle gives
$\sup_n\|\Lambda_n\|=D_{q,2}<\infty$.
If these norms are unbounded, the same principle provides a
fixed $\varphi_0\in L_{q'}(\mu)$ with
$\sup_n|\Lambda_n(\varphi_0)|=\infty$. Set $\varphi=|\varphi_0|$.
Positivity gives $\Lambda_n(\varphi)\ge|\Lambda_n(\varphi_0)|$.
The law $Y=S\sqrt{\varphi(X)}$ then has $R_{n,2}\to\infty$ along
an increasing subsequence. Each fixed-$n$ error is finite, and
the law does not depend on $n$. This completes the proof.
\end{proof}

\subsection{Nagata dimension and domination: Proofs}
\label{sec:nagata-proofs}
This subsection proves the geometric results of
Section~\ref{sec:nagata}. The geometric count below yields
Stone's domination under finite global Nagata dimension.
The subsequent constructions prove the counterexamples for
unbounded responses. The general analytic consistency
criteria in Subsection~\ref{sec:conditional-moments} do not
require these geometric assumptions.

\begin{proof}[Proof of Corollary~\ref{nagata-global-consistency}]
Fix $n$ and write $k=k_n$. Take $n+1$ independent points
$Z_0,\ldots,Z_n$ with distribution $\mu$. For each $i$, let
$\mathcal N_i$ be the indices of the $k$ nearest neighbors of
$Z_i$ among the other $n$ points. By \eqref{defhx}, almost
surely there are no distance ties at any of these queries.
For a fixed target index $j$, put
\[
I_j=\{i\ne j:j\in\mathcal N_i\}.
\]
Thus $|I_j|=D_{n,j}$ in \eqref{augmented-selection-count}.
For $i\in I_j$, consider the closed ball
$B_i=B(Z_i,\rho(Z_i,Z_j))$. Every such ball contains $Z_j$.
The Nagata property supplies a subfamily covering all centres
$Z_i$, $i\in I_j$, with multiplicity at most $\delta+1$.
Since every retained ball contains $Z_j$, there are at most
$\delta+1$ retained balls. Each contains at most $k+1$ of the
augmented sample points, including its centre and $Z_j$.
As $j\notin I_j$, it therefore contains at most $k$ of the
centres being counted. Hence
\[
D_{n,j}=|I_j|\le k(\delta+1).
\]
Taking the conditional expectation given $Z_j=z$ and using
\eqref{density-selection-count} gives
\[
f_n(z)=\frac1k\EXP[D_{n,j}\mid Z_j=z]\le\delta+1
\qquad\text{for $\mu$-almost every $z$}.
\]
The bound is uniform in $n$. The averaging identity yields
Stone's condition. The consistency conclusion follows from
Corollary~\ref{mean-response-moments} with $p=q=\infty$.
\end{proof}

For completeness, we retain the compactness proof of the
fixed-mass bound in \eqref{fixed-mass-reverse}, in the global
case used here.

\begin{proof}[Global bound for $f_u^*$]
Suppose that $\X$ is complete and separable and
$\dim_{\mathrm{Nag}}(\X)\le\delta$. Fix $z\in\X$ and
$0<u\le1$. By inner regularity of a Borel probability measure
on a complete separable metric space, it suffices to prove
$\mu(K)\le(\delta+1)u$ for an arbitrary compact subset $K$ of
\[
\{x:\mu(B(x,\rho(x,z)))\le u\}.
\]
Let $D=\{a_m:m\ge1\}$ be a countable dense subset of a
nonempty $K$. For every $N$, apply the Nagata property to
$B(a_m,\rho(a_m,z))$, $1\le m\le N$. The retained balls
cover all $a_1,\ldots,a_N$ and all contain $z$, so there are
at most $\delta+1$ of them. Repeating centres if necessary,
denote their centres by $x_1^N,\ldots,x_{\delta+1}^N\in K$.
Compactness gives a subsequence along which
$x_i^N\to x_i\in K$ for every $i$.

We claim that $B(x_i,\rho(x_i,z))$, $1\le i\le\delta+1$,
cover $K$. Their union is closed, so it suffices to show
that it contains $D$. For each fixed $m$, some index $i$
covers $a_m$ along a further infinite subsequence. Passing
to the limit in
$\rho(a_m,x_i^N)\le\rho(x_i^N,z)$ gives
$\rho(a_m,x_i)\le\rho(x_i,z)$. This proves the claim.
Since each $x_i\in K$, every limiting ball has mass at most
$u$. Thus $\mu(K)\le(\delta+1)u$ and $f_u^*(z)\le\delta+1$.
\end{proof}

\begin{proof}[Proof of Proposition~\ref{nagata-unbounded-counterexample}]
Choose $c>0$ so that
\[
\sum_{i=2}^\infty\frac{c}{i(\log i)^2}=1,
\qquad p_i=\frac{c}{i(\log i)^2}.
\]
Let $\mathcal T$ be the countable bouquet of intervals
\[
\mathcal T=\{o\}\cup\{(i,a):i\ge2,\ 0<a\le1\},
\]
with common root $o$ and tree metric
\[
\rho(o,(i,a))=a,\qquad
\rho((i,a),(j,b))=
\begin{cases}
|a-b|,&i=j,\\
a+b,&i\ne j.
\end{cases}
\]
Put no mass at $o$, and put Lebesgue measure of total mass $p_i$
on spine $i$:
\[
\mu(E)=\sum_{i=2}^\infty p_i
\int_0^1\IND_E((i,a))\,da
\qquad(E\text{ Borel}).
\]
The space is separable. It is also complete: a Cauchy sequence
whose distances from $o$ tend to zero converges to $o$; otherwise
it eventually stays on one compact spine. Every sphere meets
each spine in at most two points, so every sphere has zero
$\mu$-mass. In particular, the no-distance-ties assumption holds.

Each spine has Nagata dimension one relative to $\mathcal T$, on the
infinite scale. To verify the ambient-relative assertion, take
a finite disconnected family of balls with centres on one
spine, and order the centres by increasing distance from $o$.
Only the ball at the first centre can reach $o$: any later ball
that reaches $o$ would contain the first centre. Thus the
multiplicity outside the spine is at most one. On the spine
it is at most two, by the usual interval argument: among three
centres, a common point of their balls lies on one side of the
middle centre, forcing one of the extreme balls to contain that
centre. Consequently, $\mathcal T$ is sigma-finite dimensional in the
sense of Nagata. Its dimension in itself is infinite on every
positive scale: for any $0<a<\min\{s,1\}$, the balls
$B((i,a),a)$ form a disconnected family and all contain $o$.

For $0<u<1$, define
\[
f_u^*(y)=\frac1u\mu\{x:\mu(B(x,\rho(x,y)))\le u\},
\qquad N(u)=\#\{i\ge2:p_i\ge u\}.
\]
Fix $y=(j,a)$ with $0<a\le u/2$. For every $i\ne j$ with
$p_i\ge u$, consider centres $x=(i,t)$ with
\[
0<t\le\frac{u-a}{2p_i}.
\]
This interval is contained in $(0,1]$. Since
$\rho(x,y)=t+a$, the corresponding ball satisfies
\[
\mu(B(x,\rho(x,y)))
\le p_i(2t+a)+(1-p_i)a
=a+2p_i t\le u.
\]
The mass of these centres on spine $i$ is $(u-a)/2\ge u/4$.
Therefore
\begin{equation}
\label{porcupine-star-lower}
f_u^*(y)\ge\frac{N(u)-1}{4}
\qquad\text{if }y=(j,a),\quad 0<a\le u/2.
\end{equation}

Now put $u_n=k_n/(2n)$. Fix $x,y$ with
$t=\mu(B(x,\rho(x,y)))\le u_n$. Since spheres have zero mass,
the number $Z$ of the other $n-1$ sample points closer to $x$
than $y$ has distribution $\mathrm{Bin}(n-1,t)$. Markov's
inequality gives
\[
\PROB\{Z\ge k_n\}
\le\frac{(n-1)t}{k_n}
\le\frac{n-1}{2n}<\frac12.
\]
Thus $g_n(x,y)\ge1/2$ on this set. By \eqref{fn},
\begin{equation}
\label{porcupine-fn-lower}
f_n(y)\ge\frac{n}{2k_n}
\mu\{x:\mu(B(x,\rho(x,y)))\le u_n\}
=\frac14 f_{u_n}^*(y).
\end{equation}

Define one noiseless response by
\[
Y=m(X),\qquad m(o)=0,\qquad
m((i,a))=[\log(e/a)]^3.
\]
The response is noiseless, and for every $1\le r<\infty$,
\[
\EXP|Y|^r=\int|m|^r\,d\mu
=\int_0^1[\log(e/a)]^{3r}\,da
=e\int_1^\infty t^{3r}e^{-t}\,dt<\infty.
\]
Thus the same regression function belongs to every finite
$L_r(\mu)$. For all sufficiently large $n$, $N(u_n)\ge2$.
For these $n$, the averaging identity and
\eqref{porcupine-star-lower}--\eqref{porcupine-fn-lower} give
\begin{align}
A_n(m)
&\ge\frac{N(u_n)-1}{16}
\int_0^{u_n/2}[\log(e/a)]^3\,da\nonumber\\
&\ge\frac{(N(u_n)-1)u_n}{64}
[\log(2e/u_n)]^3.
\label{porcupine-mean-lower}
\end{align}
For the second inequality, integrate only over
$a\in[u_n/4,u_n/2]$.

To see that this bound diverges for every $u_n\to0$, choose
$j$ with $p_{j+1}<u\le p_j$. Then $N(u)=j-1$, and
\[
(N(u)-1)u[\log(2e/u)]^3
\ge(j-2)p_{j+1}[\log(2e/p_j)]^3
\asymp\log j
\longrightarrow\infty.
\]
Since $k_n/n\to0$, \eqref{porcupine-mean-lower}
shows that $A_n(m)\to\infty$. For every $1\le r<\infty$,
Jensen's inequality on the joint distribution of the sample
and the independent query yields
\[
R_{n,r}\ge
\left|\EXP\int(m_n-m)\,d\mu\right|^r
=\left|A_n(m)-\int m\,d\mu\right|^r
\longrightarrow\infty.
\]
Each fixed-$n$ risk is finite, since \eqref{averaging} gives
\[
R_{n,r}\le2^{r-1}
\left[A_n(|m|^r)+\|m\|_r^r\right]
\le2^{r-1}\left(\frac n{k_n}+1\right)\|m\|_r^r<\infty.
\]
The joint distribution of $(X,Y)$ is fixed throughout, and
the regression function depends neither on $r$ nor on the
sequence $(k_n)$.
\end{proof}

\begin{remark}[Finite dimension on a positive scale]
\label{finite-scale-unbounded}
The preceding failure also occurs on a space of finite Nagata
dimension at one fixed scale. To see this, take the following
closed subset of the countable $\ell_1$ tree, with its induced
metric:
\[
\X_0=[0,1/2]e_0\ \cup\ \bigcup_{i\ge2}[1,2]e_i,
\]
where the $e_i$ are distinct unit coordinate vectors. It is
complete and separable. Distinct interval components have
distance at least one, so balls of radius less than $1/2$
remain within one component. The interval covering property
therefore gives
$\dim_{\mathrm{Nag}}^{1/2}(\X_0,\X_0)\le1$.

Give the hub $[0,1/2]e_0$ Lebesgue density one, hence total
mass $1/2$, and give branch $[1,2]e_i$ uniform mass
$q_i=p_i/2$, with $p_i$ as in the preceding proof. Write
$\mu_0$ for this probability measure. Every sphere meets each
component in at most two points, so \eqref{defhx} holds.
Put
\[
T(v)=\sum_{i:q_i\le v}q_i.
\]
The integral test for $\sum_i[i(\log i)^2]^{-1}$ gives
$T(v)\asymp1/\log(1/v)$ as $v\downarrow0$.

For a target $z=a e_0$ with $0<a\le u/2$ and $0<u<1/2$,
every query $x=t e_i$, $1\le t\le2$, has
\[
\mu_0(B(x,\rho(x,z)))=q_i+a.
\]
The ball contains its entire branch and the hub segment
$[0,a]e_0$, and it contains no other branch. Thus every
branch with $q_i\le u/2$ contributes its full mass to the
fixed-mass reverse set, giving
\[
f_u^*(a e_0)\ge\frac{T(u/2)}u.
\]
The binomial argument in \eqref{porcupine-fn-lower} applies
unchanged with $\mu_0$, so, for $u_n=k_n/(2n)$,
\[
f_n(a e_0)\ge\frac{T(u_n/2)}{4u_n}
\qquad(0<a\le u_n/2).
\]

Define $m(a e_0)=[\log(e/a)]^3$ for $a>0$, put $m(0)=0$,
and put $m=0$ on every far branch. The noiseless response
$Y=m(X)$ has every finite moment, by the same exponential-tail
integral as above. Integrating only over
$a\in[u_n/4,u_n/2]$ yields
\[
A_n(m)\ge\frac{T(u_n/2)}{16}[\log(2e/u_n)]^3
\longrightarrow\infty.
\]
Jensen's inequality again gives $R_{n,r}\to\infty$ for every
finite $r\ge1$, for this one fixed law and every sequence with
$k_n/n\to0$. Each fixed-$n$ risk is finite.
\end{remark}

% ...


\begin{thebibliography}{99}

\bibitem[Biau and Devroye(2015)]{BiDe15}
\textsc{Biau, G.} and \textsc{Devroye, L.} (2015).
\newblock \emph{Lectures on the Nearest Neighbor Method}.
\newblock Springer, Cham.

\bibitem[C\'erou and Guyader(2006)]{cerou2006nearest}
\textsc{C\'erou, F.} and \textsc{Guyader, A.} (2006).
\newblock Nearest neighbor classification in infinite dimension.
\newblock \emph{ESAIM: Probability and Statistics} \textbf{10} 340--355.

\bibitem[Cesari and Colomboni(2020)]{CeCo21}
\textsc{Cesari, T. R.} and \textsc{Colomboni, R.} (2020).
\newblock A nearest neighbor characterization of Lebesgue points
in metric measure spaces.
\newblock \emph{Mathematical Statistics and Learning} \textbf{3}(1) 71--112.

\bibitem[Chaudhuri and Dasgupta(2014)]{ChDa14}
\textsc{Chaudhuri, K.} and \textsc{Dasgupta, S.} (2014).
\newblock Rates of convergence for nearest neighbor classification.
\newblock In \emph{Advances in Neural Information Processing
Systems 27}, pages 3437--3445.

\bibitem[Collins et~al.(2020)]{CoKuPe20}
\textsc{Collins, B.}, \textsc{Kumari, S.} and \textsc{Pestov, V.~G.} (2020).
\newblock Universal consistency of the $k$-NN rule in metric spaces
and Nagata dimension.
\newblock \emph{ESAIM: Probability and Statistics} \textbf{24} 914--934.

\bibitem[Cover and Hart(1967)]{CoHa67}
\textsc{Cover, T.~M.} and \textsc{Hart, P.~E.} (1967).
\newblock Nearest neighbor pattern classification.
\newblock \emph{IEEE Trans. Inform. Theory} \textbf{13}(1) 21--27.

\bibitem[Devroye(1982)]{Dev82}
\textsc{Devroye, L.} (1982).
\newblock Necessary and sufficient conditions for the pointwise
convergence of nearest neighbor regression function estimates.
\newblock \emph{Zeitschrift f\"ur Wahrscheinlichkeitstheorie
und verwandte Gebiete} \textbf{61} 467--481.

\bibitem[Devroye and Gy\"orfi(1985)]{DeGy85}
\textsc{Devroye, L.} and \textsc{Gy\"orfi, L.} (1985).
\newblock \emph{Nonparametric Density Estimation: The $L_1$ View}.
\newblock Wiley, New York.

\bibitem[Devroye et~al.(1994)]{DeGyLuKr94}
\textsc{Devroye, L.}, \textsc{Gy\"orfi, L.}, \textsc{Krzy\.zak, A.} and \textsc{Lugosi, G.} (1994).
\newblock On the strong universal consistency of nearest neighbor
regression function estimates.
\newblock \emph{Ann. Statist.} \textbf{22}(3) 1371--1385.

\bibitem[D\"oring et~al.(2018)]{DoGyWa18}
\textsc{D\"oring, M.}, \textsc{Gy\"orfi, L.} and \textsc{Walk, H.} (2018).
\newblock Rate of convergence of $k$-nearest-neighbor classification
rule.
\newblock \emph{Journal of Machine Learning Research} \textbf{18}(227) 1--16.

\bibitem[Ferraty and Vieu(2006)]{ferraty2006nonparametric}
\textsc{Ferraty, F.} and \textsc{Vieu, P.} (2006).
\newblock \emph{Nonparametric Functional Data Analysis:
Theory and Practice}.
\newblock Springer, New York.

\bibitem[Gy\"orfi et~al.(2002)]{GyKoKrWa02}
\textsc{Gy\"orfi, L.}, \textsc{Kohler, M.}, \textsc{Krzy\.zak, A.} and \textsc{Walk, H.} (2002).
\newblock \emph{A Distribution-Free Theory of Nonparametric Regression}.
\newblock Springer, New York.

\bibitem[Gy\"orfi et~al.(2002)]{GyScWa02}
\textsc{Gy\"orfi, L.}, \textsc{Sch\"afer, D.} and \textsc{Walk, H.} (2002).
\newblock Relative stability of global errors of nonparametric
function estimators.
\newblock \emph{IEEE Trans. Inform. Theory} \textbf{48}(8) 2230--2242.

\bibitem[Gy\"orfi and Weiss(2021)]{GyWe21}
\textsc{Gy\"orfi, L.} and \textsc{Weiss, R.} (2021).
\newblock Universal consistency and rates of convergence of
multiclass prototype algorithms in metric spaces.
\newblock \emph{Journal of Machine Learning Research} \textbf{22}(151) 1--25.

\bibitem[Hanneke et~al.(2021)]{HaKoSaWe21}
\textsc{Hanneke, S.}, \textsc{Kontorovich, A.}, \textsc{Sabato, S.} and \textsc{Weiss, R.} (2021).
\newblock Universal Bayes consistency in metric spaces.
\newblock \emph{Ann. Statist.} \textbf{49}(4) 2129--2150.
\newblock See also the supplementary material, Lemma~A.1.

\bibitem[Kumari and Pestov(2024)]{KuPe24}
\textsc{Kumari, S.} and \textsc{Pestov, V.~G.} (2024).
\newblock Universal consistency of the $k$-NN rule in metric spaces
and Nagata dimension. II.
\newblock \emph{ESAIM: Probability and Statistics} \textbf{28} 132--160.

\bibitem[Nagata(1963)]{nagata1}
\textsc{Nagata, J.} (1963).
\newblock On a special metric characterizing a metric space of
$\dim\le n$.
\newblock \emph{Proceedings of the Japan Academy} \textbf{39} 278--282.

\bibitem[Ostrand(1965)]{ostrand}
\textsc{Ostrand, P.~A.} (1965).
\newblock A conjecture of J. Nagata on dimension and metrization.
\newblock \emph{Bulletin of the American Mathematical Society} \textbf{71} 623--625.

\bibitem[Pestov(2026)]{Pes26}
\textsc{Pestov, V.~G.} (2026).
\newblock Universal consistency of the $k$-NN rule in metric spaces
and Nagata dimension. III.
\newblock arXiv:2512.17058v3. Version of June 4, 2026.
\newblock \url{https://arxiv.org/abs/2512.17058v3}.

\bibitem[Preiss(1983)]{preiss83}
\textsc{Preiss, D.} (1983).
\newblock Dimension of metrics and differentiation of measures.
\newblock In \emph{General Topology and its Relations to Modern
Analysis and Algebra V (Prague, 1981)}, pages 565--568.
Sigma Series in Pure Mathematics, 3. Heldermann, Berlin.

\bibitem[Preiss et~al.(2021)]{PrRiTi21}
\textsc{Preiss, D.}, \textsc{Riss, E.} and \textsc{Ti\v{s}er, J.} (2021).
\newblock A set of positive Gaussian measure with uniformly zero
density everywhere.
\newblock \emph{Journal of the European Mathematical Society} \textbf{23}(7) 2439--2466.

\bibitem[Stone(1977)]{Sto77}
\textsc{Stone, C.~J.} (1977).
\newblock Consistent nonparametric regression.
\newblock \emph{Ann. Statist.} \textbf{5}(4) 595--620.



\end{thebibliography}
\end{document}